\documentclass[sigconf]{acmart}
\renewcommand\footnotetextcopyrightpermission[1]{}

\usepackage{stfloats}
\usepackage{tabularx}
\usepackage{subcaption}
\usepackage{longtable}
\usepackage{lipsum}
\usepackage{algorithm}
\usepackage{algpseudocode}
\usepackage{amsmath}
\usepackage{threeparttable}
\usepackage{stfloats}
\usepackage{float}

\AtBeginDocument{
  \providecommand\BibTeX{{
    \normalfont B\kern-0.5em{\scshape i\kern-0.25em b}\kern-0.8em\TeX}}}

\begin{document}
\title{Transparent Early ICU Mortality Prediction with Clinical Transformer and Per-Case Modality Attribution}

\author{Alexander Bakumenko}
\authornote{Author to whom correspondence should be addressed.}
\affiliation{
  \institution{Clemson University}
  \city{Charleston, SC}
  \country{USA}}
\email{abakume@clemson.edu}
\orcid{0000-0001-7212-9573}

\author{Janine Hoelscher}
\affiliation{
  \institution{Clemson University}
  \city{Clemson, SC}
  \country{USA}
}
\email{janineh@clemson.edu}

\author{Hudson Smith}
\affiliation{
  \institution{Clemson University}
  \city{Clemson, SC}
  \country{USA}
}
\email{dane2@clemson.edu}

\begin{abstract}
Early identification of intensive care patients at risk of in-hospital mortality enables timely intervention and efficient resource allocation. Despite high predictive performance, existing machine learning approaches lack transparency and robustness, limiting clinical adoption. We present a lightweight, transparent multimodal ensemble that fuses physiological time-series measurements with unstructured clinical notes from the first 48 hours of an ICU stay. A logistic regression model combines predictions from two modality-specific models: a bidirectional LSTM for vitals and a finetuned ClinicalModernBERT transformer for notes. This traceable architecture allows for multilevel interpretability: feature attributions within each modality and direct per-case modality attributions quantifying how vitals and notes influence each decision. On the MIMIC-III benchmark, our late-fusion ensemble improves discrimination over the best single model (AUPRC 0.565 vs. 0.526; AUROC 0.891 vs. 0.876) while maintaining well-calibrated predictions. The system remains robust through a calibrated fallback when a modality is missing. These results demonstrate competitive performance with reliable, auditable risk estimates and transparent, predictable operation, which together are crucial for clinical use.
\end{abstract}


\maketitle
\pagestyle{plain}

\section{Introduction}
Early ICU mortality risk estimation is valuable only if its outputs can be trusted and acted upon. Clinicians need probabilities they can confidently threshold, case-level rationales they can check, and modeling logic they can understand. Multimodal architectures may deliver strong prediction performance, but complex models are difficult to audit and govern at the bedside. Our aim is a solution that keeps these practical promises.

\begin{figure*}[]
    \centering
    \includegraphics[width=1\linewidth]{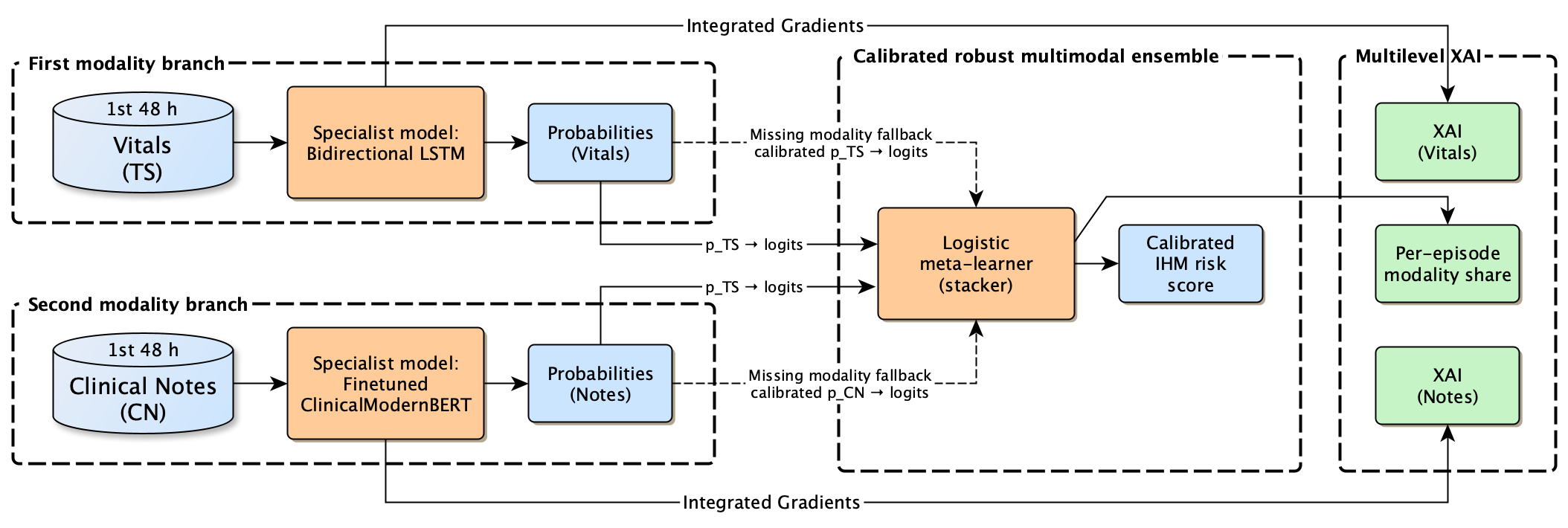}
    \caption{Transparent multimodal ensemble for early ICU mortality prediction. Two specialist models, a bidirectional LSTM for time-series vitals (TS), and a finetuned ClinicalModernBERT transformer for clinical notes (CN), produce probability outputs that are converted to standardized logits. A logistic regression model (meta-learner) combines these logits to generate a calibrated in-hospital mortality (IHM) risk score. Orange blocks indicate modular, independently trainable components. The system provides multilevel explainable AI (XAI): Integrated Gradients attributions identify influential vitals variables/time-steps and note tokens within each specialist, while per-case modality shares quantify how vitals and notes influence a decision. When a modality is unavailable, the system falls back to the calibrated probability, ensuring graceful degradation.}
    \label{fig:ICU_risk_main}
\end{figure*}

\subsection{Clinical Motivation}
The Intensive Care Unit (ICU) is a critical hub for the treatment of patients with life-threatening conditions and is associated with substantial mortality rates and high resource utilization \cite{vincent2006sepsis}. Clinicians are under immense pressure to make high-stakes, rapid decisions based on a continuous flow of complex patient data. Early and accurate identification of patients with the highest risk of in-hospital mortality (IHM) is of great importance. Such predictions can enable timely, targeted interventions, facilitate more effective allocation of limited resources (e.g., specialist staff, equipment), and improve communication with patients and families regarding prognosis \cite{topol2019high}. Consequently, developing robust and reliable decision support for this task remains a primary objective in computational medicine.
In practice, an early-risk model would surface a prioritized list of patients on the ICU dashboard (for instance, after the first 48 hr), with service-defined thresholds prompting a nurse or rapid-response review rather than a hard alarm to limit alert fatigue. The display can pair each risk estimate with a brief rationale (e.g., whether physiologic trends or note content drive the signal), supporting accountable actions such as intensified monitoring, escalation to intensivist consult, targeted resource planning, or earlier goals-of-care discussions.

\subsection{The Challenge}

The trajectory of a patient in the ICU is captured through a combination of structured and unstructured data. Structured signals (physiological time-series and labs) provide objective, quantitative measurements \cite{rajkomar2018scalable}, while clinical notes capture clinical reasoning, intent, and qualitative observations. Although combining these modalities showed an improvement in predictive performance \cite{khadanga2019using}, it presents significant technical challenges. Many state-of-the-art multimodal models rely on complex fusion architectures (such as intricate attention-based transformers) that are difficult to interpret and govern. Limited transparency remains a barrier to clinical adoption \cite{adadi2018peeking}. 

In high-stakes settings like ICU, transparency and calibration primarily serve bedside decisions and underpin safety and trust. 
Following established frameworks on explainability for clinical applications and trustworthy AI \cite{combi2022manifesto, tonekaboni2019clinicians, eu2019ethics}, we operationalize transparency as interpretability (case-level rationales), understandability (inspectable fusion mechanisms), complemented by calibration and robustness to missing data \cite{bakumenko2025clinictransparentoperabledesign}. Black-box multimodal fusion cannot meet these requirements. Interpretable and auditable designs are recommended for high-stakes decisions \cite{rudin2019stop,tonekaboni2019clinicians}.
Despite progress in multimodal prediction, clinical adoption struggles with opaque modeling, unaddressed reliability and operation under real-data conditions \cite{tonekaboni2019clinicians}.

This motivates our probability-level late fusion with a linear meta-learner and per-case modality attribution, coupled with within-branch attributions and explicit calibration.
Desirable systems should therefore be accurate, interpretable at both feature and modality levels, well-calibrated, and robust when a modality is weak or absent, which in practice means graceful degradation with a defined fallback for the missing modality rather than catastrophic failure. We treat missing-modality robustness as a first-class requirement and evaluate it explicitly.

Feature-level interpretability means tracing a prediction to specific inputs (e.g., vital-sign variables at particular hours or salient tokens/phrases in the notes) with signed attributions that indicate risk-increasing vs. risk-reducing evidence (e.g., via Integrated Gradients or SHAP) \cite{sundararajan2017axiomatic,lundberg2017unified}. Modality-level interpretability means explaining, per case (episode), how much each data source (vitals vs. notes) pulled the final risk up or down \cite{bakumenko2025clinictransparentoperabledesign}. In our linear stacking, this is an exact decomposition of the meta-decision into weighted logit contributions from each branch.
\subsection{Our Contribution}

We introduce a transparent and lightweight multimodal architecture for early ICU mortality prediction on MIMIC-III. It offers multilevel (feature level and modality level per episode) interpretability, reliable predictions, and robustness to missing data, leveraging standardized pipelines for vitals \cite{harutyunyan2019multitask} and clinical notes \cite{khadanga2019using}. 
Figure \ref{fig:ICU_risk_main} summarizes the proposed architecture and its transparency properties.

Our contributions are threefold:
\begin{enumerate}
    \item \textbf{Lightweight, modular late-stage fusion.} 
    We propose probability-level stacking via logistic regression on standardized logits from independently trained specialists (LSTM for vitals, finetuned ClinicalModernBERT for clinical notes). This yields quantified discrimination gains over the strongest standalone branch while keeping each branch plug-and-play: either can be monitored and upgraded without retraining the other or altering downstream calibration.
    \item \textbf{Multilevel interpretability.} 
    We provide feature-level per-case explanations within each specialist (which vitals/time-steps or note tokens drive risk) and per-case modality attribution at the fusion layer (how much vitals versus notes contributed). This enables drill-down interpretability from case-level predictions to contributing features.
    \item \textbf{Calibration and missing-modality robustness.} 
    We quantify reliability (probability calibration) to see how well predicted risks match observed in-hospital mortality frequencies. We also evaluate deterministic missing-modality scenarios. The proposed multimodal architecture is well-calibrated by design and yields graceful degradation under missing-modality conditions.
\end{enumerate}

\section{Background and Related Work}
 \label{sec:related}
\subsection{ICU Mortality Prediction}
Early ICU risk stratification has traditionally relied on physiology-based severity scores (APACHE II, SAPS II, SOFA), which summarize multi–organ dysfunction into calibrated mortality risks \cite{knaus1985apacheii,legall1993sapsii,vincent1996sofa}. The availability of large, de-identified critical-care datasets such as MIMIC-III and the newer MIMIC-IV has catalyzed reproducible research and benchmarking for in-hospital mortality prediction from high-resolution ICU time-series, labs, and free-text notes \cite{johnson2016mimiciii,johnson2023mimiciv,harutyunyan2019multitask}. 
 
Recent competitive approaches span models handling informative missingness and irregular ICU time series (GRU-D; GRU-ODE-Bayes \cite{che2018recurrent,de2019gru}), attention/transformer families for sparse clinical sequences (SAnD; self-/semi-supervised transformers \cite{song2018attend,tipirneni2022self}), long-context clinical language models that exploit full notes (Clinical-Longformer/BigBird; NYUTron, GatorTron \cite{li2022clinical,jiang2023health,yang2022gatortron}), and multimodal transformers that jointly attend over structured EHR and notes \cite{lyu2023multimodal}. On MIMIC-III IHM, results typically fall within AUROC $\approx$ 0.87–0.90 and AUPRC $\approx$ 0.53–0.62, varying with horizon, cohort filters, and splits (see \cite{lyu2023multimodal,tipirneni2022self,battula2024enhancing}). Despite strong performance, end-to-end deep fusion models are computationally heavy, offer limited modality-level transparency, and rarely report calibration or robustness to missing modalities.
We position our work as a lightweight, calibrated late-fusion alternative targeting this accuracy band while maintaining auditability and graceful degradation when a modality is missing.
\subsection{Multimodal Fusion in Healthcare and beyond}
Multimodal ML addresses three coupled problems: 
(i) representation-learning feature representations for each modality and optionally shared cross-modal representations;
(ii) alignment-handling asynchrony, missingness, and cross-rate signals; and 
(iii) fusion-combining information at different stages \cite{baltrusaitis2019multimodal,ramachandram2017deep}. 
Canonical regimes include early fusion (feature concatenation under temporal/structural alignment), intermediate fusion (shared latent spaces or cross-modal attention/transformers for information exchange) \cite{ngiam2011multimodal,tsai2019multimodal}, and late fusion (ensembles/stacking over per-modality predictions) \cite{wolpert1992stacked,breiman1996stacked}. These choices trade flexibility and accuracy against compute, data demands, and auditability. 

In clinical deployments, late fusion via stacking is more compelling: specialist models can be trained, validated, and monitored independently, then combined by a lightweight meta-learner on out-of-sample predictions. This decoupling (i) eases engineering and re-validation when a branch evolves, (ii) supports graceful degradation under missing or degraded modalities (the ensemble falls back to available specialists), and (iii) exposes modality-level contributions that facilitate governance. For text, domain-adapted transformer encoders trained on clinical corpora provide strong document representations and complementary signal to physiology \cite{alsentzer2019publicly,singhal2023large}.
We implement probability-to-logit stacking using a linear logistic meta-learner trained on standardized logits from specialists trained over the first 48 hr of the ICU stay. This preserves interpretability with minimal computational overhead while retaining competitive discrimination performance.

\subsection{Explainable AI and Calibration in Medicine}
 Limited transparency and miscalibration are key barriers to deploying ML in critical care \cite{adadi2018peeking,rudin2019stop,tonekaboni2019clinicians}. 

At the feature level within each specialist, Integrated Gradients (IG) provides fine-grained feature attributions (saliency scores over input variables or tokens) with formal guarantees of sensitivity and implementation invariance \cite{sundararajan2017axiomatic,simonyan2013deep}.
Alternative paradigms include SHAP (Shapley values from cooperative game theory satisfying local accuracy/consistency) and Layer-wise Relevance Propagation (LRP, which redistributes output scores backward under relevance-conservation rules) \cite{lundberg2017unified,bach2015lrp}. Our fusion layer is a logistic meta-learner whose per-sample contributions are coefficient-weighted logits. Within branches (LSTM for vitals and transformer for notes), we use IG for architecture-agnostic attributions, whereas LRP needs architecture-specific rules, and SHAP is costly for long sequences.
\par{\textbf{Multilevel explanations.}} Beyond feature-level saliency, linear stacking permits an exact per-episode modality attribution by decomposing the meta-decision into contributions from each specialist, answering the governance-relevant question “which source drove this prediction?”. When non-linear meta-learners are used, SHAP provides a consistent alternative at the fusion layer \cite{lundberg2017unified}.
\par{\textbf{Calibration.}} Clinically, action thresholds depend on calibration-predicted probabilities matching observed outcome frequencies. Standard metrics include Expected Calibration Error, Brier score, and reliability diagrams. Post-hoc methods (Platt scaling, isotonic regression) can correct miscalibrated outputs \cite{niculescu2005predicting,guo2017calibration,brier1950verification}. 

\section{Materials and Methods}
\label{sec:methods}

\subsection{Dataset and Cohort Definition}
\label{sec:dataset}
\subsubsection{MIMIC-III Overview}
We use the publicly available MIMIC-III ICU database and adopt the standardized benchmark configuration and task definitions introduced by Harutyunyan et al.~\cite{harutyunyan2019multitask} for in-hospital mortality (IHM) prediction using the first 48 hr of ICU measurements. The benchmark specifies how to construct instances for IHM and provides a consistent test split to enable reproducible evaluation and comparison. For the text modality, we follow the preprocessing pipeline of Khadanga et al.~\cite{khadanga2019using}, which defines how to collect and prepare clinical notes within the same prediction window.

\subsubsection{Inclusion/Exclusion Criteria}
Following these pipelines, we include ICU stays with at least 48 hr of data (the IHM label is predicted at $t{=}48$ hr) and use the benchmark’s predefined test set with a validation split from the remaining data. For the notes modality, we exclude clinical notes without an associated chart time and exclude patients with no notes during the first 48 hr. These rules mirror the original implementations for structured vitals and unstructured notes.

\subsubsection{Cohort Statistics}
After aligning the two modalities (first 48 hr vitals and at least one note within 0--48 hr), the final cohort sizes are: train $N{=}10{,}978$, validation $N{=}2{,}435$, and test $N{=}2{,}448$. During note matching, missing or unloadable notes were observed and excluded (train: 361 episodes; validation: 89; test: 78). Each retained episode contains a 48 hr multivariate time series with 76 features per hour (see vitals representation below) and a concatenated 0--48 hr notes document.
The task is imbalanced, with in-hospital mortality prevalence of $\approx11\%$ on the held-out test split (similar across splits and consistent with prior MIMIC-III reports \cite{harutyunyan2019multitask}). Accordingly, we emphasize AUPRC as the primary metric along with other standard metrics and report calibration analyses.

\subsection{Data Pre-processing}

\subsubsection{Time-Series Vitals}
We follow the Harutyunyan et al.~\cite{harutyunyan2019multitask} pipeline for structured time-series vitals (TS). Time series are resampled into hourly intervals; when multiple measurements occur within an hour, the last value is used. Missing values are imputed using last observation carried forward when available, otherwise with a pre-specified “normal” value. A binary observation mask is created per variable to indicate whether a value was observed versus imputed. Numeric inputs are standardized per variable after imputation. Concatenating the 17 variable-wise masks and values yields a 76-dimensional vector per hour; the IHM instance uses the first 48 such hourly vectors.

The 17 clinical variables are: capillary refill rate, diastolic blood pressure, fraction of inspired oxygen, Glasgow Coma Scale (eye, motor, verbal, and total), glucose, heart rate, height, mean arterial pressure, oxygen saturation, respiratory rate, systolic blood pressure, temperature, weight, and pH. Harutyunyan et al.~\cite{harutyunyan2019multitask} also report the "normal" values used during imputation for each variable (e.g., HR = 86, SBP = 118) and whether each is modeled as continuous or categorical. 

\subsubsection{Clinical Notes}
For the unstructured modality, we replicate the MIMIC-III clinical notes (CN) processing described by Khadanga et al.~\cite{khadanga2019using}. For the IHM task, all notes charted within the first 48 hr of ICU stay are concatenated into a single document $N_T$ used at $t{=}48$ hr. Notes without a valid chart time are removed, and ICU stays without any notes are excluded. This matches our inclusion/exclusion criteria and explains the drop in sample size when merging modalities. 
\subsection{Multimodal Ensemble Architecture}
\subsubsection{Specialist Models}

To enable flexible and interpretable multimodal fusion, we independently train several specialist models on different representations of ICU patient data: structured time-series variables and unstructured clinical notes. Each model is optimized as a standalone binary classifier for in-hospital mortality prediction, producing a probability per episode. Their predicted probability scores are used as inputs to our late-fusion ensemble.
We train specialists across multiple architectures for each modality. For time-series vitals, we implement 1D convolutional neural network (CNN), recurrent neural network (RNN), and bidirectional Long Short-Term Memory (LSTM) models \cite{bai2018empirical,hochreiter1997long,lipton2015learning}. For clinical notes, we use 1D CNN on tokenized text, frozen transformer embeddings using ModernBERT and ClinicalModernBERT, and a fully finetuned ClinicalModernBERT model following prior work \cite{vaswani2017attention,alsentzer2019publicly}.

\paragraph{Time-Series Based Models.}
We implement a two-layer recurrent neural network using a Bidirectional LSTM (8 units) followed by a unidirectional LSTM (16 units), with dropout layers in between to mitigate overfitting. The input is a 48-hour multivariate sequence of normalized clinical variables. A final dense sigmoid layer outputs mortality risk. The model is trained using binary cross-entropy loss and optimized with the Adam optimizer.
As an alternative to recurrent architectures, we design a convolutional model that applies two stacked 1D convolutional layers (32 and 64 filters) with ReLU activations, followed by global average pooling. This configuration allows the model to detect temporal patterns in the physiological signals over short time windows. Like the LSTM, it is trained using Adam and binary cross-entropy.
We include a lightweight stacked RNN model with the same hidden sizes as the LSTM to provide a non-gated recurrent baseline. This model provides insight into whether more complex gating mechanisms yield performance gains.

\paragraph{Clinical Notes Based Models.}
For free-text clinical notes, we tokenize and pad the first 48-hour concatenated text for each patient. A 1D CNN with a filter size of 5 and 128 filters is applied to word embeddings (learned during training), followed by max pooling, dropout, and a dense layer. This model captures local n-gram patterns relevant to mortality prediction 
and represents a modification of the original model by Khadanga et al. \cite{khadanga2019using}.
We assess the predictive power of pretrained general ModernBERT and domain-specific ClinicalModernBERT representations by using their [CLS] token embeddings for each patient’s 48-hour ICU clinical notes. Each episode is tokenized into a long-context sequence (up to 8192 tokens), encoded with the transformer, and the [CLS] vector from the last hidden layer is extracted and L2-normalized. These document-level embeddings are then standardized and used to train a logistic regression classifier for in-hospital mortality prediction, following the embedding approach of \cite{bakumenko2025advancing}. No finetuning is performed for these embedding-based models, isolating the predictive value of frozen representations.
We also finetune a transformer-based classifier built on a ClinicalModernBERT model. The entire model, including all encoder layers, is finetuned end-to-end on our mortality prediction task using training data and cross-entropy loss. Sequences are truncated or padded to a maximum of 8192 tokens.

Each model is evaluated independently on a held-out test set, and probability predictions are stored for downstream fusion and explanation.

\subsubsection{Ensemble Method}
    
In our late-fusion ensemble, specialist models independently generate IHM probability scores, which are aggregated by a meta-learner for the final prediction.

We consider several meta-learners to aggregate these base-model outputs, including logistic regression (LOGREG), random forests (RF), gradient-boosted trees (GBM), XGBoost (XGB), multilayer perceptrons (MLP), and simple averaging (AVG). In our implementation:
\begin{itemize}
    \item \textbf{LOGREG} and \textbf{MLP} operate on base-model \emph{logits}:
    \begin{equation}
    z_i = \log\!\left(\frac{p_i}{1-p_i}\right),
    \label{eq:logit_def}
    \end{equation}
    where $p_i$ is the probability from specialist $i$, with clipping at $10^{-6}$.
    These logits are standardized on the validation set.
    \item \textbf{Tree-based models} (RF, GBM, XGB) receive the raw probabilities directly, since their split criteria are invariant to monotonic transformations such as logits or scaling.
    \item \textbf{AVG} computes the simple mean of the raw probabilities, assigning equal weight to all included specialists.
\end{itemize}

For the LOGREG meta-learner, standardized base-model logits $z_i$ are linearly combined as
\begin{equation}
\text{logit}(p_{\text{meta}}) = b_\text{eff} + \sum_i w_i z_i,
\label{eq:meta_logit}
\end{equation}
where $b_\text{eff}$ is the effective intercept and $w_i$ are the learned coefficients.

The meta-learner is trained using validation-set predictions and labels, and final performance is evaluated on a held-out test set using independently generated test predictions.

Model contributions to the ensemble prediction are quantified via the learned coefficients (in LOGREG), normalized to the sum of absolute values, or via feature importances (in tree-based models). For simple averaging, contributions are uniformly weighted. While model-level weights remain global in our framework, we also explore per-sample contribution attribution for linear models by decomposing the final prediction into weighted input contributions normalized across the ensemble inputs.

\subsection{Explainability Framework}

Our explainability strategy operates at two levels: (i) \emph{model-specific feature attributions}, which explain the inner workings of each specialist branch (time-series vitals and clinical notes), and (ii) \emph{ensemble-level modality attribution}, which explains how the fusion meta-learner combines these branches on a per-patient basis.

\subsubsection{Model-Specific Feature Attributions}

\paragraph{Vitals (LSTM)}
For the recurrent time-series model, we explain predictions using Integrated Gradients (IG) targeting the probability of the positive class $p(y=1)$ from the network’s sigmoid output. The baseline $x_0$ is the all-zero sequence, consistent with the model’s Masking (\textit{mask\_value=0}) layer, representing the absence of measurements. We integrate along the linear path $x_\alpha = x_0 + \alpha(x-x_0)$ with $20$ steps:
\begin{equation}
\text{IG}(x_i) \approx (x_i - x_{0,i}) \cdot \frac{1}{S}\sum_{s=1}^{S} 
\frac{\partial p(x_{s})}{\partial x_i}, \quad S = 20.
\label{eq:integrated_gradients}
\end{equation}
Attributions are aggregated for one-hot encoded features to produce feature-level explanations, and categorical values are decoded back to canonical labels.

Outputs include: (i) printed ranked tables of the most important attributions by $|\text{saliency}|$, (ii) top-K positive (risk-increasing) and top-K negative (risk-reducing) drivers, (iii) top features ranked by mean $|\text{saliency}|$, and (iv) contiguous multi-hour “windows” of positive/negative saliency with the observed values at their peak. A thresholded heatmap (global top 10\% of $|\text{saliency}|$) provides a sparse, interpretable visualization across hours $\times$ variables.
Clinically, these outputs highlight not just isolated hours but sustained episodes of abnormal physiology (e.g., prolonged low GCS motor scores, sustained hypotension).

\paragraph{Clinical notes (finetuned ClinicalModernBERT)}
For free-text clinical notes, we use Integrated Gradients on token embeddings of the finetuned ClinicalModernBERT model, targeting the positive-class probability. The baseline is the PAD embedding sequence, representing absence of content. We interpolate over $25$ steps, backpropagating gradients to obtain token-level attributions.

To ensure readability, we (i) merge subword pieces into surface words, (ii) collapse frequent duplicates while retaining the peak attribution, and (iii) form negation-aware bigrams (e.g., ``denies pain'', ``no insulin'') by merging common negators with the following term. Stopwords are pruned conservatively with a whitelist for clinical abbreviations (e.g., NSR, HR, BP), which can be further revisited and refined based on the underlying data. We then apply sign-separated thresholds (keep $\geq 20\%$ of top magnitude for each sign), and present two short ranked lists: risk-increasing and risk-reducing drivers, each with peak attribution values. For the strongest positive and negative terms, a compact context snippet (bounded by sentence punctuation and capped at $\sim$200 characters) is extracted to show local text evidence. 

Clinically, these outputs surface language markers that the model associates with higher or lower mortality risk (e.g., “poor prognosis” or “opacification” as risk-increasing, versus “nsr” or “transfer to step-down” as risk-reducing). The design emphasizes compactness, enabling clinicians to directly inspect influential terms.

\subsubsection{Ensemble-Level Modality Attribution}

At the ensemble level, we interpret how the LOGREG meta-learner (Equation \eqref{eq:meta_logit}) combines the standardized logits from the vitals and notes branches. 
For each patient, we compute per-modality contributions as
$c_i = w_i z_i$,
and normalized shares
$\tilde c_i = \frac{|c_i|}{\sum_j |c_j|}$,
which quantify how strongly each modality "pulled" the ensemble logit. 

The explanation reports 
branch votes (specialist probabilities), 
ensemble probability with predicted class, 
the linear decision equation in logit units, and the dominant modality contribution with percentage share. Agreement labels ("agree high", "agree low", "conflict") summarize whether branches concur or conflict. This makes each meta-decision auditable: clinicians see which branch pulled the risk estimate upward or downward and by how much.

By combining ensemble-level attribution with branch-specific IG explanations, we provide a multilevel interpretability pipeline. Clinicians can first identify whether vitals or notes dominated the ensemble’s decision, and then drill down within that branch to see which time-points, variables, or note phrases drove the risk estimate. This supports both global model governance and per-patient case (episode) review.

\subsection{Experimental Setup}

All models were trained on the benchmarked MIMIC-III splits with the official predefined test set held out for final evaluation. Validation predictions were used both to train the fusion meta-learner and to fit post-hoc calibration mappings, while test predictions were reserved strictly for reporting.

Class discrimination was evaluated using area under the precision–recall curve (AUPRC) as the primary metric, with ROC AUC, F1, accuracy, precision, recall, and balanced accuracy as secondary measures. Thresholded metrics were computed at a fixed cutoff of 0.5. 
We selected specialist models based on highest AUPRC within each modality. For the ensemble meta-learner, we selected based on highest AUPRC combined with the ability to provide direct per-case modality attribution. Interpretability was examined at two levels: per-case feature attributions within each specialist and per-case modality contributions.

Calibration was quantified using the Brier score, Expected Calibration Error (ECE) with 20 equal-frequency bins, calibration slope and intercept from logistic recalibration, and reliability diagrams. Post-hoc calibration was applied independently to each specialist branch using Platt scaling, temperature scaling, or isotonic regression, with the method chosen on the validation set by lowest ECE. These calibrated probabilities were then applied to test data.

Robustness was evaluated by simulating deterministic missing-modality scenarios at inference. When a modality was missing, the system used its calibrated probability alongside probability of the available modality.

\section{Results}
Our late-fusion ensemble (LSTM + ft\_ClinicalModernBERT, 
logistic regression meta-learner) 
on the held-out test set described in Section \ref{sec:dataset}
achieves an AUPRC of 0.565 [0.503, 0.625], outperforming any standalone specialist (best notes: 0.526; best vitals: 0.485; 
see Table~\ref{tab:discrimination}). We selected this configuration for its balance of discrimination, simplicity, and interpretability,
and it remains well-calibrated (Table~\ref{tab:calibration_pre_post}).
Unless otherwise noted, we report mean performance with 95\% confidence intervals (mean [95\% CI]) estimated via 1,000 bootstrap resamples of the test set.

\begin{table}[t]
\centering
\footnotesize
\caption{Model performance summary (AUC and AUPRC) for standalone specialists and selected ensemble on the test set. Results shown as mean [95\% CI]. Best mean value per column in bold.}
\label{tab:discrimination}
\begin{tabular}{lcc}
\toprule
\textbf{Model} & \textbf{AUC} & \textbf{AUPRC} \\
\midrule
\multicolumn{3}{l}{\textit{Clinical Notes}} \\
ft\_ClinicalModernBERT  & 0.876 [0.856, 0.895] & 0.526 [0.462, 0.586] \\
cn\_CNN                 & 0.823 [0.797, 0.849] & 0.450 [0.392, 0.514] \\
emb\_ClinicalModernBERT & 0.831 [0.807, 0.856] & 0.420 [0.359, 0.480] \\
emb\_ModernBERT         & 0.721 [0.691, 0.751] & 0.239 [0.200, 0.282] \\
\midrule
\multicolumn{3}{l}{\textit{Time-series Vitals}} \\
LSTM                    & 0.855 [0.832, 0.877] & 0.485 [0.425, 0.545] \\
ts\_CNN                 & 0.829 [0.802, 0.854] & 0.417 [0.359, 0.479] \\
RNN                     & 0.836 [0.811, 0.861] & 0.413 [0.355, 0.473] \\
\midrule
\multicolumn{3}{l}{\textit{Selected Ensemble (LogReg)}} \\
LSTM + ft\_ClinicalModernBERT  & \textbf{0.891 [0.872, 0.909]} & \textbf{0.565 [0.503, 0.625]} \\
\bottomrule
\end{tabular}
\end{table}
\label{sec:results}

\subsection{Performance of Specialist Models}

Among clinical notes models, AUPRC was highest for the fully finetuned ClinicalModernBERT (ft\_ClinicalModernBERT, 0.526 [0.462, 0.586]), followed by the cn\_CNN, trained directly on tokenized clinical text (0.450 [0.392, 0.514]). ClinicalModernBERT embeddings with a logistic regression classifier emb\_ClinicalModernBERT achieved an AUPRC of 0.420 [0.359, 0.480], while ModernBERT embeddings performed lower at 0.239 [0.200, 0.282]. 
The emb\_Modern-BERT and emb\_ClinicalModernBERT models used frozen (precomputed) transformer embeddings as input features for a logistic regression classifier \cite{bakumenko2025advancing}, whereas the ft\_ClinicalModernBERT model was finetuned with its own classification head, trained end-to-end on the classification task. All ModernBERT-based models utilized full long-context input windows of 8192 tokens to maximize context usage.

\begin{figure}[b]
    \centering
    \includegraphics[width=1\linewidth]{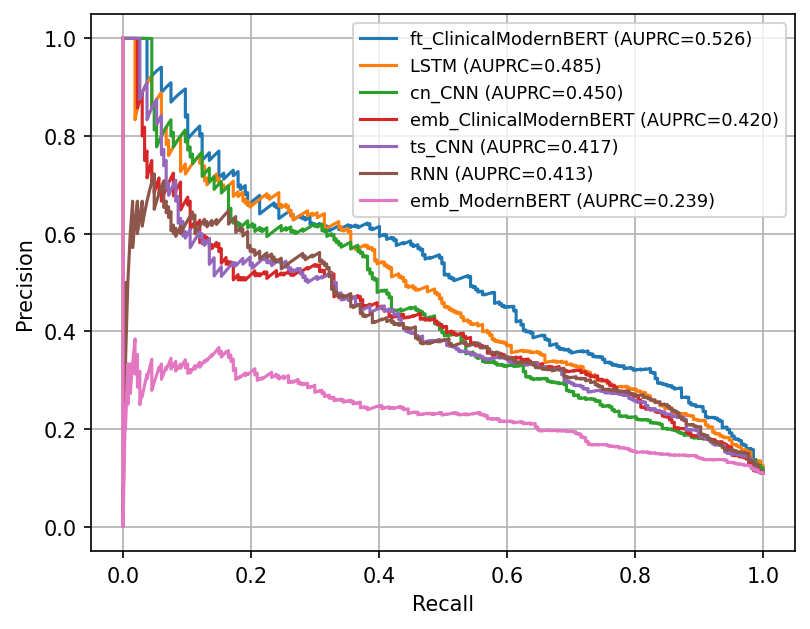}
    \caption{Precision–Recall Curve for all specialist classifiers.}
    \label{fig:auprc_curves}
\end{figure}

For time-series vitals, the bidirectional LSTM achieved the highest AUPRC among vitals models at 0.485 [0.425, 0.545], followed by the ts\_CNN and RNN. 

Overall, the finetuned ClinicalModernBERT (notes) and the LSTM (vitals) demonstrated the highest AUPRC scores and were selected for ensemble fusion. Figure~\ref{fig:auprc_curves} shows precision-recall curves for all specialist models. Complete performance metrics including threshold-dependent measures are provided in Appendix Table~\ref{tab:metric_summary_full}. Confidence intervals were estimated via 1,000 bootstrap resamples of the test set.

\vspace{10pt}
\subsection{Performance of Ensemble Models}

\begin{table}[b]
\centering
\footnotesize
\caption{Ensemble performance (AUC and AUPRC) for all meta-learner and specialist pairing combinations. Best within each pairing in bold.}
\label{tab:ensemble_summary}
\begin{tabular}{
    p{0.37\linewidth}
    >{\centering\arraybackslash}p{0.25\linewidth}
    >{\centering\arraybackslash}p{0.25\linewidth}
}
\hline
\textbf{Algorithm} & \textbf{AUC} & \textbf{AUPRC} \\
\hline
\addlinespace[2pt]
\multicolumn{3}{l}{\textit{LSTM + ft\_ClinicalModernBERT}} \\
\addlinespace[2pt]
AVG    & \textbf{0.893 [0.875, 0.911]} & 0.562 [0.499, 0.623] \\
LOGREG & 0.891 [0.872, 0.909] & \textbf{0.565 [0.503, 0.625]} \\
GBM    & 0.879 [0.860, 0.898] & 0.480 [0.417, 0.542] \\
RF     & 0.861 [0.837, 0.882] & 0.490 [0.432, 0.553] \\
MLP    & 0.884 [0.864, 0.902] & 0.535 [0.470, 0.595] \\
XGB    & 0.876 [0.856, 0.896] & 0.527 [0.468, 0.586] \\
\hline
\addlinespace[2pt]
\multicolumn{3}{l}{\emph{LSTM + cn\_CNN}} \\
\addlinespace[2pt]
AVG    & 0.869 [0.847, 0.890] & \textbf{0.508 [0.447, 0.570]} \\
LOGREG & \textbf{0.870 [0.849, 0.891]} & 0.506 [0.444, 0.565] \\
GBM    & 0.866 [0.845, 0.887] & 0.474 [0.414, 0.537] \\
RF     & 0.843 [0.818, 0.867] & 0.456 [0.395, 0.518] \\
MLP    & 0.863 [0.841, 0.883] & 0.490 [0.428, 0.549] \\
XGB    & 0.861 [0.838, 0.882] & 0.495 [0.435, 0.556] \\
\hline
\addlinespace[2pt]
\multicolumn{3}{l}{\emph{ts\_CNN + ft\_ClinicalModernBERT}} \\
\addlinespace[2pt]
AVG    & 0.884 [0.865, 0.902] & 0.528 [0.470, 0.589] \\
LOGREG & \textbf{0.886 [0.867, 0.903]} & \textbf{0.539 [0.480, 0.601]} \\
GBM    & 0.876 [0.858, 0.894] & 0.493 [0.432, 0.551] \\
RF     & 0.846 [0.824, 0.868] & 0.449 [0.389, 0.508] \\
MLP    & 0.880 [0.860, 0.898] & 0.515 [0.456, 0.575] \\
XGB    & 0.864 [0.843, 0.884] & 0.491 [0.434, 0.547] \\
\hline
\addlinespace[2pt]
\multicolumn{3}{l}{\emph{ts\_CNN + cn\_CNN}} \\
\addlinespace[2pt]
AVG    & \textbf{0.858 [0.836, 0.879]} & \textbf{0.477 [0.419, 0.537]} \\
LOGREG & 0.857 [0.836, 0.878] & 0.465 [0.406, 0.525] \\
GBM    & 0.855 [0.834, 0.876] & 0.469 [0.408, 0.527] \\
RF     & 0.815 [0.791, 0.841] & 0.420 [0.364, 0.476] \\
MLP    & 0.850 [0.828, 0.871] & 0.453 [0.394, 0.511] \\
XGB    & 0.843 [0.819, 0.866] & 0.463 [0.405, 0.523] \\
\hline
\end{tabular}
\end{table}

\begin{table*}[b]
\footnotesize
\caption{
Pre- and post-calibration on the test set (mean [95\% CI]) for TS (LSTM), CN (ft\_ClinicalModernBERT), and the best-performing ensemble. Discrimination (AUC/AUPRC) shown for reference.
}
\label{tab:calibration_pre_post}
\begin{tabular}{lcccccc}
\toprule
\textbf{Model} & \textbf{AUC} & \textbf{AUPRC} & \textbf{Brier} & \textbf{ECE} & \textbf{Slope} & \textbf{Intercept} \\
\midrule
TS(LSTM) (pre) & 0.855 [0.832,0.877] & 0.485 [0.425,0.545] & 0.075 [0.068,0.083] & 0.145 [0.138,0.155] & 0.873 [0.785,0.973] & $-0.326$ [$-0.509$,$-0.145$] \\
TS(LSTM) (post) & 0.854 [0.832,0.877] & 0.456 [0.399,0.515] & 0.075 [0.068,0.083] & 0.144 [0.137,0.154] & 0.827 [0.696,0.962] & $-0.364$ [$-0.616$,$-0.132$] \\
CN(ft\_ClinicalModernBERT) (pre) & 0.876 [0.856,0.895] & 0.526 [0.462,0.586] & 0.105 [0.098,0.112] & 0.209 [0.198,0.219] & 0.921 [0.831,1.022] & $-1.468$ [$-1.620$,$-1.319$] \\
CN(ft\_ClinicalModernBERT) (post) & 0.873 [0.853,0.892] & 0.473 [0.412,0.529] & 0.072 [0.065,0.078] & 0.143 [0.136,0.152] & 1.003 [0.886,1.135] & $-0.119$ [$-0.324$,0.094] \\
Ensemble (pre) & 0.891 [0.872,0.909] & 0.565 [0.503,0.625] & 0.067 [0.060,0.074] & 0.133 [0.126,0.142] & 0.982 [0.883,1.088] & $-0.162$ [$-0.347$,0.024] \\
Ensemble (post) & 0.891 [0.872,0.909] & 0.565 [0.503,0.625] & 0.067 [0.060,0.074] & 0.133 [0.126,0.142] & 0.979 [0.880,1.084] & $-0.166$ [$-0.351$,0.020] \\
\bottomrule
\end{tabular}
\end{table*}

\begin{figure}[ht]
    \raggedleft
    \includegraphics[width=0.9\linewidth]{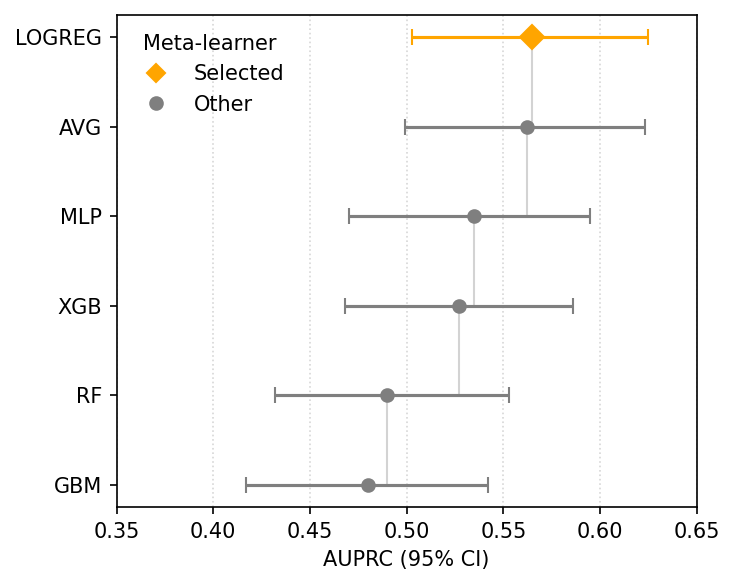}
    \caption{AUPRC discrimination (mean [95\% CI]) comparison of six meta-learning algorithms for the LSTM + ft\_ClinicalModernBERT pairing.}
    \label{fig:AUPRC_ensembles}
\end{figure}

Table~\ref{tab:ensemble_summary} summarizes discrimination performance of all evaluated late-fusion configurations across six meta-learning strategies and four specialist pairings. Among the tested configurations, the combination of the LSTM time-series model with the finetuned ClinicalModernBERT notes model consistently achieved the strongest results. Within this pairing, the logistic regression meta-learner yielded the highest AUPRC (0.565 [0.503, 0.625]), which is especially important for imbalanced classification tasks. While some alternative meta-learners (e.g., averaging or boosted trees) approached similar AUC levels, their precision-recall tradeoffs were inferior (Figure~\ref{fig:AUPRC_ensembles}).

We selected the LSTM + ft\_ClinicalModernBERT ensemble with logistic regression for its balance of discrimination, simplicity, and interpretability (the logistic meta-learner's coefficients provide direct modality-level weights, enabling per-case attribution).
Complete performance metrics, including threshold-dependent measures, are provided in Appendix Table~\ref{tab:ab_fusion_split}.

Figure~\ref{fig:AUPRC_specialist_ensemble} visualizes this result by directly comparing all standalone specialists against the selected ensemble, demonstrating that the ensemble provides a consistent AUPRC boost over individual models. This illustrates the benefit of integrating complementary temporal (vitals) and textual (notes) modalities through late fusion.

\begin{figure}[ht]
    \centering
    \includegraphics[width=1\linewidth]{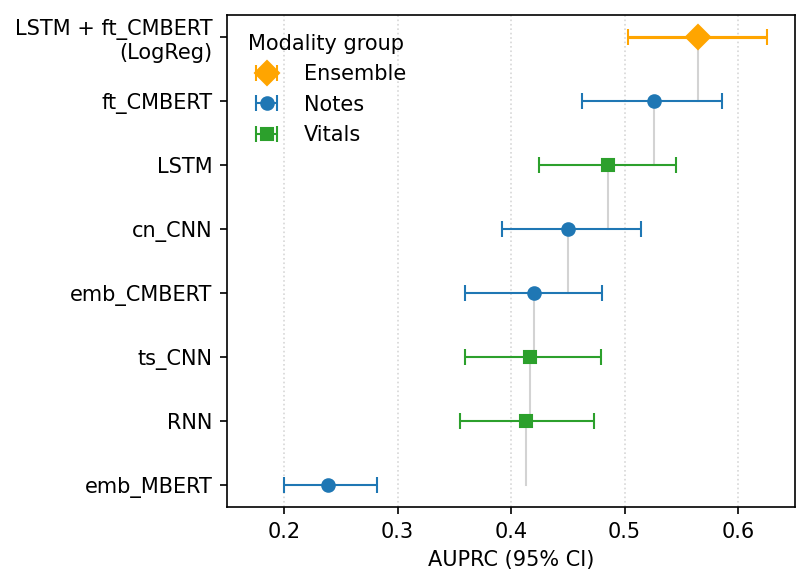}
    \caption{AUPRC discrimination (mean [95\% CI]) for all specialist models and the selected ensemble on the test set. Models emb\_ModernBERT, emb\_ClinicalModernBERT, and ft\_ClinicalModernBERT denoted as emb\_MBERT, emb\_CMBERT, and ft\_CMBERT respectively.}
    \label{fig:AUPRC_specialist_ensemble}
\end{figure}

\subsection{Calibration and Reliability}
\begin{figure*}[b]
  \centering
  \begin{minipage}{0.32\textwidth}
    \centering
    \includegraphics[width=\linewidth]{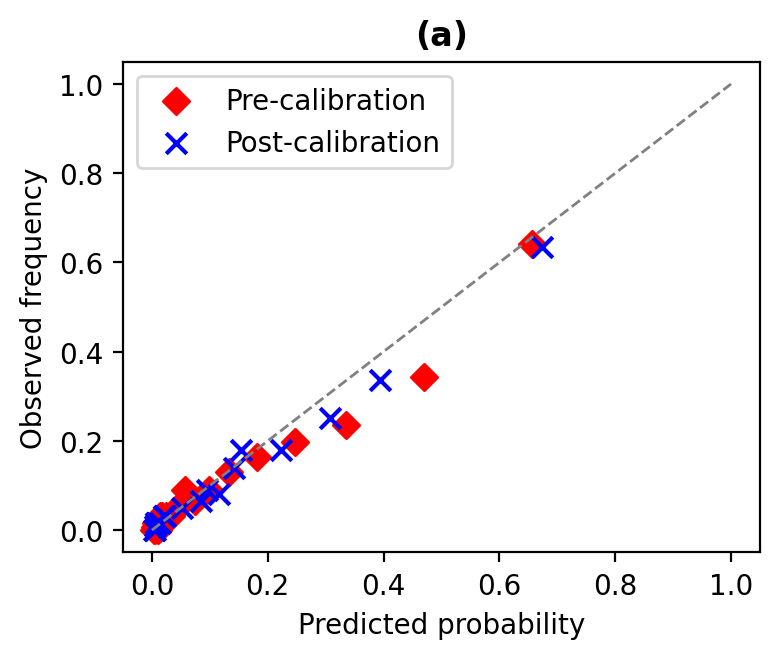}
  \end{minipage}\hfill
  \begin{minipage}{0.32\textwidth}
    \centering
    \includegraphics[width=\linewidth]{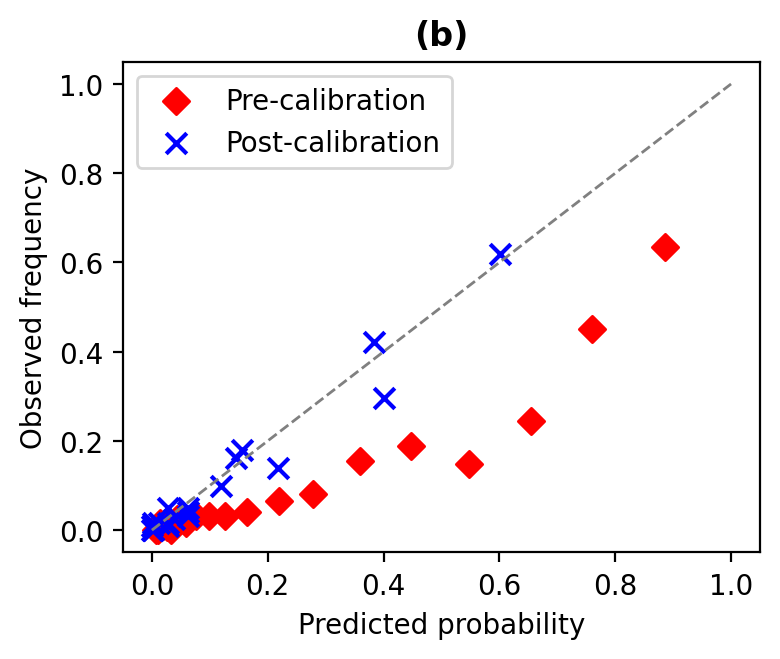}
  \end{minipage}\hfill
  \begin{minipage}{0.32\textwidth}
    \centering
    \includegraphics[width=\linewidth]{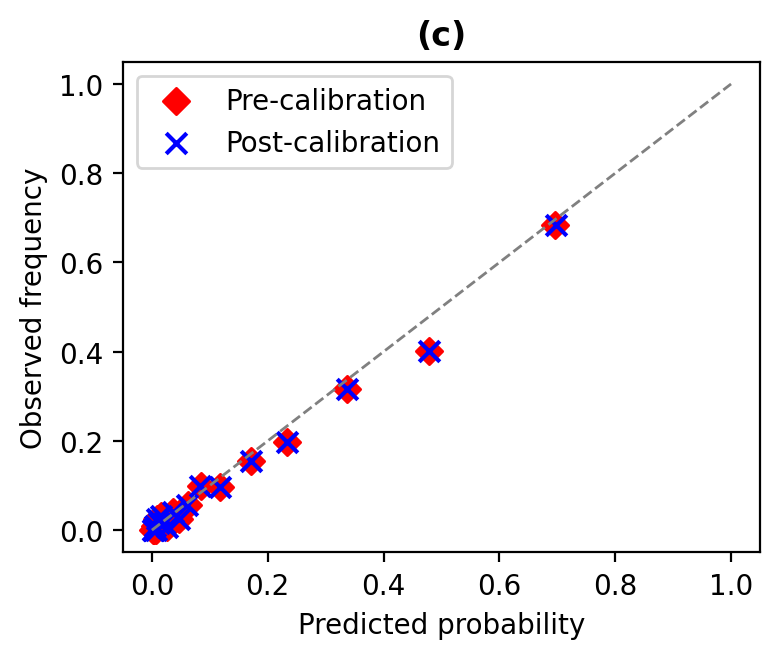}
  \end{minipage}
  \vspace{-0.7em}
  \caption{Reliability diagrams before and after isotonic regression for (a) the LSTM vitals specialist, (b) the ft\_ClinicalModernBERT notes specialist, and (c) the logistic stacker ensemble. Each panel shows pre-calibration (raw probabilities) and post-calibration (isotonic regression) curves; points closer to the diagonal indicate better calibration.}
  \label{fig:reliability_pre_post}
\end{figure*}

Table~\ref{tab:calibration_pre_post} shows calibration metrics for both specialists and the ensemble.
Both specialists selected isotonic regression as the best calibrator on the validation set.
On the held-out test set, the notes branch (ft\_ClinicalModernBERT)
shows substantial reliability gains after calibration.
Adaptive-ECE and Brier loss both decrease notably (see Table~\ref{tab:calibration_pre_post}), and the recalibration slope moves to 1.003 [0.886, 1.135]. These changes reflect sizable improvements in calibration and bring the slope very close to its ideal value of 1.
The pre-calibration diagram shows clear over-confidence at higher scores; post-cal better aligns with the diagonal (Figure~\ref{fig:reliability_pre_post}b).

For the LSTM vitals branch, calibration barely changes the curve (Adaptive-ECE 0.145 to 0.144; Brier unchanged at 0.075). Isotonic slightly flattens extremes (slope 0.873 to 0.827; Figure~\ref{fig:reliability_pre_post}a).

The logistic stacker is well calibrated: Adaptive-ECE 0.133 [0.126, 0.142], Brier 0.067 [0.060, 0.074], slope 0.982 [0.883, 1.088], intercept -0.162 [-0.347, 0.024]. Post-hoc mapping does not change these metrics, as expected for a log-loss-trained stacker (Figure~\ref{fig:reliability_pre_post}c).

\subsection{Per-Episode Modality Attribution and Disagreement}

We summarize per-case modality contributions via the distribution of ensemble shares.
Shares are mirror-symmetric by construction (TS \(=1-\)CN). The CN median is \(\approx 0.37\) (TS \(\approx 0.63\)), indicating a modest tilt toward vitals overall with the wide spread: roughly half the cohort lies in a moderate CN band (\(0.25{\text{–}}0.45\)), while near-balanced cases (\(0.45{\text{–}}0.55\)) are less common. Strong dominance is rare (TS share \(>0.80\) or CN share \(>0.80\)), showing that, in most episodes, both modalities contribute.
This pattern is visualized in Appendix Figure~\ref{fig:hist_of_share}.

\begin{table*}[t]
\centering
\caption{Missing-modality fallback on the test set (mean [95\% CI]). Fallback uses the post-hoc calibrated probability of the missing branch alongside the output of the available branch.}
\label{tab:missing_modality}
\begin{tabular}{lcccc}
\toprule
\textbf{Scenario} & \textbf{AUC} & \textbf{AUPRC} & \textbf{Brier} & \textbf{ECE} \\
\midrule
Both-present (ensemble, pre) & 0.891 [0.872,0.909] & 0.565 [0.503,0.625] & 0.067 [0.060,0.074] & 0.133 [0.126,0.142] \\
Notes-absent (TS cal)        & 0.854 [0.832,0.877] & 0.456 [0.399,0.515] & 0.075 [0.068,0.083] & 0.144 [0.137,0.154] \\
Vitals-absent (CN cal)       & 0.873 [0.853,0.892] & 0.473 [0.412,0.529] & 0.072 [0.065,0.078] & 0.143 [0.136,0.152] \\
\bottomrule
\end{tabular}
\end{table*}

In Appendix Figure~\ref{fig:scatter}, each episode is placed by the vitals logit (x) and notes logit (y), with color showing the ensemble probability. 
We denote risk-agreement patterns using arrows on the specialists’ logits: "↓↓" for both low-risk (Agree low) logits, "↑↑" for both high-risk (Agree high) logits, and "↑↓" for disagreement (Conflict).
The dense lower-left quadrant (↓↓) captures the majority where both branches indicate low risk (mean \(p_{\text{meta}}=0.044\)). The small upper-right cluster (↑↑) shows rare high-risk agreement (mean \(p_{\text{meta}}=0.674\)). In conflicts, when notes are high but vitals low (upper-left, \(N=360\)), the ensemble assigns intermediate risk (mean \(p_{\text{meta}}=0.306\)); when vitals are high but notes low (lower-right, \(N=34\)), it is slightly higher (mean \(p_{\text{meta}}=0.365\)), consistent with a modestly larger meta-weight on vitals. 

\begin{figure}[ht]
    \centering
    \includegraphics[width=\linewidth]{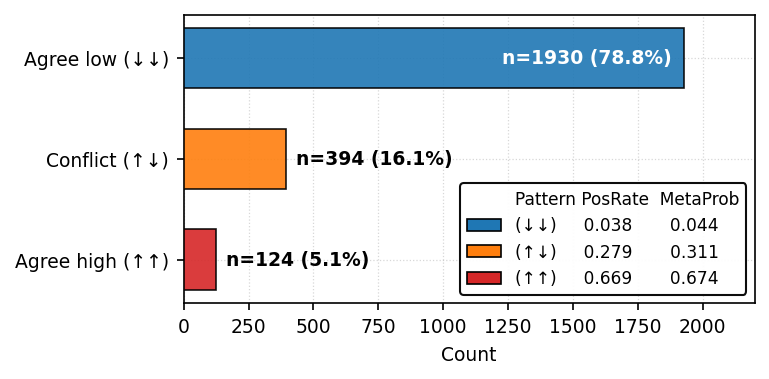}
    \caption{Agreement patterns of assigned risk on the test set by specialist logit signs. Bars show distribution of Agree low (↓↓), Conflict (↑↓), and Agree high (↑↑) categories with counts and prevalence (\%). Legend shows observed event rates (PosRate) and mean ensemble probabilities (MetaProb).}
    \label{fig:agr_counts}
\end{figure}

Figure~\ref{fig:agr_counts} partitions the cohort by base-branch agreement. Most episodes are \emph{Agree low} (↓↓; 78.8\%) with a low observed event rate (PosRate 0.038) and low mean ensemble probability (\(0.044\)). 
\emph{Conflicts} (↑↓) are 16.1\% with a higher event rate (0.279) and intermediate mean ensemble probability (\(0.311\)), identifying the slice where per-episode shares guide clinicians to the driver (TS vs.\ CN) for drill-down. \emph{Agree high} (↑↑) is rare (5.1\%), aligns with uncommon high-risk episodes (PosRate 0.669), and receives high ensemble probabilities (mean \(0.674\)). 

Overall, the ensemble boosts agreement and dampens disagreement. Conflicts, having markedly higher event rates, are the prime targets for per-episode explanations.

\subsection{Robustness to Missing/Degraded Modalities}

Table~\ref{tab:missing_modality} shows performance under deterministic missing-modality scenarios at inference. When notes are absent, AUPRC drops to 0.456 [0.399, 0.515], a decrease of 0.109 from the both-present ensemble; when vitals are absent, AUPRC drops to 0.473 [0.412, 0.529], a decrease of 0.092. AUC remains above 0.85 in both cases (notes-absent: 0.854 [0.832, 0.877]; vitals-absent: 0.873 [0.853, 0.892]). 
ECE and Brier remain stable across scenarios: ECE ranges from 0.133 (ensemble) to 0.144 (both fallbacks), and Brier from 0.067 (ensemble) to 0.075 (notes-absent).

\section{Discussion}
\label{sec:discussion}
\subsection{Analysis of late fusion}
The best-performing specialists were the LSTM for vitals and the finetuned ClinicalModernBERT for clinical notes. Fusing their outputs with a logistic meta-learner improved discrimination over either branch alone, reaching AUPRC 0.565 [0.503, 0.625] and AUC 0.891 [0.872, 0.909].
This gain is consistent with complementary errors across modalities:
when both agree on high (or low) risk, the ensemble confidently follows; when they disagree, the ensemble assigns intermediate risk rather than favoring one source.
Agreement analysis shows most episodes are agree-low with very low event rates, a small agree-high slice carries high risk, and about one sixth of cases are true conflicts enriched for events. Modality shares are centered modestly toward vitals (median notes share $\approx$ 0.37), but each modality is decisive for a substantial subset.

\subsection{Calibration and reliability}
Calibration results separate clearly from discrimination. For notes, isotonic regression significantly improved reliability on test, reducing Adaptive-ECE from 0.209 [0.198, 0.219] to 0.143 [0.136, 0.152] and Brier from 0.105 [0.098, 0.112] to 0.072 [0.065, 0.078], with slope and intercept near ideal. The vitals branch was already reasonable, and post-hoc mapping produced negligible changes (ECE 0.145 to 0.144; Brier unchanged at 0.075). The logistic stacker trained on standardized logits was well-calibrated out of the box (a Platt mapping did not change its metrics). 

\subsection{Interpretability and governance}
The system provides multilevel explanations that address both feature-level interpretability (which inputs drove risk) and modality-level interpretability (which data source dominated the decision. At the fusion layer, per-episode modality contributions quantify how vitals versus notes influenced each prediction, directing clinicians to the relevant branch for drill-down in conflict cases. Within branches, Integrated Gradients surface specific temporal patterns in vitals (e.g., sustained hypotension) and salient phrases in notes (e.g., "poor prognosis"), providing auditable evidence for case review. This design operationalizes the understandability requirement: the linear meta-learner's weights are directly inspectable, branch-specific models are independently governable, and per-case attributions enable clinicians to verify whether the system's reasoning aligns with domain knowledge before acting on its output.

\subsection{Practical considerations and limitations}
The modular design enables independent branch maintenance and graceful degradation under missing-modality conditions, addressing the robustness requirement. When either modality is unavailable, the system falls back to a calibrated single-specialist prediction rather than failing or producing uncalibrated outputs. Performance degrades measurably but not catastrophically, with AUPRC reductions of 0.09-0.11 while maintaining calibration metrics near branch-level values.

Limitations include the retrospective, single-center evaluation on MIMIC-III and the focus on the two strongest specialists for primary results. External validation is needed to assess generalizability across institutions, patient populations, and clinical workflows. Broader ablations with weaker branches would clarify the conditions under which fusion remains beneficial. In deployment, drift monitoring
would be required to maintain reliability as patient populations and documentation practices evolve.

\section{Conclusion and Future Work}
\label{sec:conclusion}
We presented a multimodal ensemble for early ICU mortality prediction that addresses key requirements for clinically credible AI ~\cite{bakumenko2025clinictransparentoperabledesign}: 
prediction performance (AUPRC 0.565, outperforming single-modality baselines), 
interpretability (per-case feature and modality attributions), 
understandability (traceable system design),
calibration (ensemble ECE 0.133 with near-ideal slope/intercept), and 
robustness (graceful degradation with calibrated fallbacks when a modality is missing). 
The architecture combines a bidirectional LSTM for time-series vitals and a finetuned ClinicalModernBERT for clinical notes via logistic regression, enabling independent branch training, transparent fusion weights, and per-episode modality contributions that clinicians can audit. On the MIMIC-III benchmark, this approach delivers competitive discrimination while maintaining operational transparency and reliability under real-data conditions.

Future work could explore validating and extending these properties in settings that mirror deployment. First, external validation on other ICU datasets would assess generalizability of both predictive performance and the explainability framework. Second, evaluating at earlier decision horizons (for example, 6, 12, and 24 hours) would clarify how modality shifts when data accumulates. Third, robustness evaluation can be broadened beyond deterministic modality absence to include variability in clinical notes length and coverage, distribution shifts in clinical language, and the ratio of imputed versus observed vitals. These analyses can be paired with targeted recalibration schedules and drift monitors based on slope/intercept and ECE trends. Finally, integrating per-case uncertainty estimates would enable selective abstention, allowing the system to defer high-uncertainty predictions to clinicians. These directions aim to convert a performant and interpretable research prototype into a clinically credible tool.

\bibliographystyle{ACM-Reference-Format}

\clearpage
\onecolumn
\appendix
\renewcommand{\thetable}{A\arabic{table}}
\renewcommand{\thefigure}{A\arabic{figure}}
\setcounter{table}{0}
\setcounter{figure}{0}

\section*{Appendix}
\addcontentsline{toc}{section}{Appendix}

\begin{table*}[ht]
\centering
\footnotesize
\begin{threeparttable}
\caption{Complete performance metrics for standalone specialist models on the test set. Results shown as mean [95\% CI]. Highest AUC and AUPRC metrics within each modality in bold. Models emb\_ModernBERT, emb\_ClinicalModernBERT, and ft\_ClinicalModernBERT denoted as emb\_MBERT, emb\_CMBERT, and ft\_CMBERT, respectively.}

\label{tab:metric_summary_full}
\begin{tabular}{lccccccc}
\hline
\textbf{Model} & \textbf{AUC} & \textbf{AUPRC} & \textit{F1}$^{\dagger}$ & \textit{Precision}$^{\dagger}$ & \textit{Recall}$^{\dagger}$ & \textit{Accuracy}$^{\dagger}$ & \textit{Balanced Acc}$^{\dagger}$ \\
\hline
\addlinespace[2pt]
\multicolumn{8}{l}{\textit{Clinical Notes}} \\
\addlinespace[1pt]
cn\_CNN                 & 0.823 [0.797,0.849] & 0.450 [0.392,0.514] & 0.448 [0.388,0.506] & 0.900 [0.888,0.913] & 0.561 [0.489,0.641] & 0.374 [0.314,0.432] & 0.669 [0.639,0.698] \\
emb\_MBERT         & 0.721 [0.691,0.751] & 0.239 [0.200,0.282] & 0.226 [0.171,0.278] & 0.873 [0.860,0.887] & 0.340 [0.261,0.424] & 0.170 [0.125,0.213] & 0.565 [0.542,0.586] \\
emb\_CMBERT & 0.831 [0.807,0.856] & 0.420 [0.359,0.480] & 0.397 [0.343,0.449] & 0.891 [0.879,0.903] & 0.500 [0.426,0.573] & 0.330 [0.277,0.385] & 0.645 [0.618,0.673] \\
ft\_CMBERT  & \textbf{0.876 [0.856,0.895]} & \textbf{0.526 [0.462,0.586]} & 0.482 [0.439,0.526] & 0.841 [0.827,0.855] & 0.374 [0.332,0.417] & 0.679 [0.620,0.735] & 0.770 [0.741,0.800] \\
\hline
\addlinespace[2pt]
\multicolumn{8}{l}{\textit{Time-series Vitals}} \\
\addlinespace[1pt]
ts\_CNN                 & 0.829 [0.802,0.854] & 0.417 [0.359,0.479] & 0.400 [0.341,0.454] & 0.890 [0.878,0.903] & 0.497 [0.429,0.570] & 0.335 [0.279,0.387] & 0.647 [0.617,0.674] \\
RNN                     & 0.836 [0.811,0.861] & 0.413 [0.355,0.473] & 0.305 [0.246,0.364] & 0.896 [0.884,0.909] & 0.560 [0.467,0.653] & 0.210 [0.163,0.261] & 0.595 [0.572,0.621] \\
LSTM           & \textbf{0.855 [0.832,0.877]} & \textbf{0.485 [0.425,0.545]} & 0.448 [0.386,0.508] & 0.904 [0.892,0.916] & 0.601 [0.523,0.674] & 0.357 [0.298,0.417] & 0.664 [0.634,0.694] \\
\hline
\end{tabular}
\begin{tablenotes}\footnotesize
\item[$\dagger$] Threshold-dependent complementary metrics; computed at a fixed threshold of 0.5.
\end{tablenotes}
\end{threeparttable}
\end{table*}

\begin{table*}[ht]
\centering
\footnotesize
\begin{threeparttable}
\caption{Complete performance metrics for late-fusion ensemble configurations. Results shown as mean [95\% CI]. Highest AUC and AUPRC within each specialist pairing in bold.}
\label{tab:ab_fusion_split}

\begin{tabular}{lccccccc}
\hline
\textbf{Algorithm} & \textbf{AUC} & \textbf{AUPRC} & \textit{F1}$^{\dagger}$ & \textit{Accuracy}$^{\dagger}$ & \textit{Precision}$^{\dagger}$ & \textit{Recall}$^{\dagger}$ & \textit{Balanced Acc}$^{\dagger}$ \\
\hline
\addlinespace[2pt]
\multicolumn{8}{l}{\textit{LSTM + ft\_ClinicalModernBERT}} \\
\addlinespace[2pt]
AVG    & \textbf{0.893 [0.875, 0.911]} & 0.562 [0.499, 0.623] & 0.532 [0.482, 0.580] & 0.900 [0.888, 0.913] & 0.546 [0.485, 0.607] & 0.519 [0.462, 0.578] & 0.733 [0.704, 0.762] \\
LOGREG & 0.891 [0.872, 0.909] & \textbf{0.565 [0.503, 0.625]} & 0.472 [0.413, 0.529] & 0.907 [0.895, 0.917] & 0.615 [0.544, 0.686] & 0.384 [0.327, 0.441] & 0.677 [0.648, 0.706] \\
GBM    & 0.879 [0.860, 0.898] & 0.480 [0.417, 0.542] & 0.416 [0.356, 0.466] & 0.896 [0.884, 0.908] & 0.536 [0.457, 0.609] & 0.341 [0.283, 0.394] & 0.652 [0.623, 0.679] \\
RF     & 0.861 [0.837, 0.882] & 0.490 [0.432, 0.553] & 0.459 [0.405, 0.513] & 0.901 [0.888, 0.913] & 0.567 [0.497, 0.641] & 0.387 [0.330, 0.440] & 0.675 [0.647, 0.702] \\
MLP    & 0.884 [0.864, 0.902] & 0.535 [0.470, 0.595] & 0.079 [0.040, 0.125] & 0.895 [0.882, 0.908] & 0.915 [0.700, 1.000] & 0.042 [0.021, 0.067] & 0.521 [0.510, 0.533] \\
XGB    & 0.876 [0.856, 0.896] & 0.527 [0.468, 0.586] & 0.447 [0.387, 0.504] & 0.904 [0.892, 0.916] & 0.602 [0.528, 0.677] & 0.357 [0.298, 0.415] & 0.664 [0.635, 0.694] \\
\hline
\addlinespace[2pt]
\multicolumn{8}{l}{\emph{LSTM + cn\_CNN}} \\
\addlinespace[2pt]
AVG    & 0.869 [0.847, 0.890] & \textbf{0.508 [0.447, 0.570]} & 0.389 [0.323, 0.451] & 0.905 [0.893, 0.917] & 0.654 [0.561, 0.741] & 0.278 [0.223, 0.330] & 0.630 [0.603, 0.656] \\
LOGREG & \textbf{0.870 [0.849, 0.891]} & 0.506 [0.444, 0.565] & 0.425 [0.363, 0.483] & 0.905 [0.893, 0.916] & 0.622 [0.540, 0.699] & 0.323 [0.266, 0.380] & 0.649 [0.621, 0.678] \\
GBM    & 0.866 [0.845, 0.887] & 0.474 [0.414, 0.537] & 0.347 [0.288, 0.409] & 0.898 [0.885, 0.910] & 0.572 [0.485, 0.664] & 0.250 [0.201, 0.305] & 0.614 [0.589, 0.640] \\
RF     & 0.843 [0.818, 0.867] & 0.456 [0.395, 0.518] & 0.422 [0.365, 0.477] & 0.896 [0.884, 0.908] & 0.540 [0.462, 0.616] & 0.348 [0.292, 0.401] & 0.656 [0.628, 0.683] \\
MLP    & 0.863 [0.841, 0.883] & 0.490 [0.428, 0.549] & 0.249 [0.188, 0.311] & 0.902 [0.889, 0.914] & 0.752 [0.636, 0.865] & 0.150 [0.109, 0.194] & 0.572 [0.552, 0.594] \\
XGB    & 0.861 [0.838, 0.882] & 0.495 [0.435, 0.556] & 0.406 [0.343, 0.464] & 0.905 [0.892, 0.916] & 0.631 [0.548, 0.715] & 0.300 [0.245, 0.355] & 0.639 [0.612, 0.667] \\
\hline
\addlinespace[2pt]
\multicolumn{8}{l}{\emph{ts\_CNN + ft\_ClinicalModernBERT}} \\
\addlinespace[2pt]
AVG    & 0.884 [0.865, 0.902] & 0.528 [0.470, 0.589] & 0.505 [0.454, 0.552] & 0.887 [0.875, 0.900] & 0.482 [0.430, 0.542] & 0.531 [0.473, 0.587] & 0.731 [0.701, 0.758] \\
LOGREG & \textbf{0.886 [0.867, 0.903]} & \textbf{0.539 [0.480, 0.601]} & 0.452 [0.394, 0.512] & 0.908 [0.895, 0.920] & 0.642 [0.562, 0.716] & 0.350 [0.294, 0.406] & 0.663 [0.635, 0.691] \\
GBM    & 0.876 [0.858, 0.894] & 0.493 [0.432, 0.551] & 0.383 [0.322, 0.443] & 0.897 [0.885, 0.910] & 0.557 [0.473, 0.634] & 0.293 [0.240, 0.348] & 0.632 [0.605, 0.659] \\
RF     & 0.846 [0.824, 0.868] & 0.449 [0.389, 0.508] & 0.407 [0.351, 0.461] & 0.891 [0.879, 0.904] & 0.503 [0.434, 0.577] & 0.342 [0.290, 0.400] & 0.651 [0.624, 0.679] \\
MLP    & 0.880 [0.860, 0.898] & 0.515 [0.456, 0.575] & 0.191 [0.134, 0.251] & 0.900 [0.887, 0.912] & 0.806 [0.667, 0.933] & 0.108 [0.073, 0.146] & 0.553 [0.535, 0.572] \\
XGB    & 0.864 [0.843, 0.884] & 0.491 [0.434, 0.547] & 0.373 [0.315, 0.430] & 0.897 [0.885, 0.910] & 0.555 [0.477, 0.637] & 0.281 [0.230, 0.333] & 0.627 [0.601, 0.654] \\
\hline
\addlinespace[2pt]
\multicolumn{8}{l}{\emph{ts\_CNN + cn\_CNN}} \\
\addlinespace[2pt]
AVG    & \textbf{0.858 [0.836, 0.879]} & \textbf{0.477 [0.419, 0.537]} & 0.390 [0.332, 0.449] & 0.901 [0.890, 0.914] & 0.600 [0.520, 0.680] & 0.289 [0.237, 0.339] & 0.633 [0.607, 0.658] \\
LOGREG & 0.857 [0.836, 0.878] & 0.465 [0.406, 0.525] & 0.324 [0.263, 0.386] & 0.899 [0.887, 0.912] & 0.607 [0.511, 0.702] & 0.222 [0.173, 0.271] & 0.602 [0.578, 0.627] \\
GBM    & 0.855 [0.834, 0.876] & 0.469 [0.408, 0.527] & 0.384 [0.326, 0.448] & 0.902 [0.890, 0.914] & 0.609 [0.521, 0.696] & 0.282 [0.234, 0.336] & 0.630 [0.606, 0.658] \\
RF     & 0.815 [0.791, 0.841] & 0.420 [0.364, 0.476] & 0.335 [0.279, 0.394] & 0.888 [0.875, 0.902] & 0.479 [0.395, 0.559] & 0.258 [0.208, 0.311] & 0.612 [0.586, 0.640] \\
MLP    & 0.850 [0.828, 0.871] & 0.453 [0.394, 0.511] & 0.199 [0.135, 0.261] & 0.898 [0.885, 0.912] & 0.704 [0.568, 0.833] & 0.116 [0.076, 0.157] & 0.555 [0.536, 0.576] \\
XGB    & 0.843 [0.819, 0.866] & 0.463 [0.405, 0.523] & 0.363 [0.304, 0.425] & 0.898 [0.887, 0.911] & 0.572 [0.484, 0.653] & 0.266 [0.215, 0.321] & 0.621 [0.596, 0.648] \\
\hline
\end{tabular}

\begin{tablenotes}\footnotesize
\item[$\dagger$] Threshold-dependent complementary metrics; computed at a fixed threshold of 0.5.
\end{tablenotes}
\end{threeparttable}
\end{table*}

\begin{figure}[ht]
    \centering
    \includegraphics[width=0.5\linewidth]{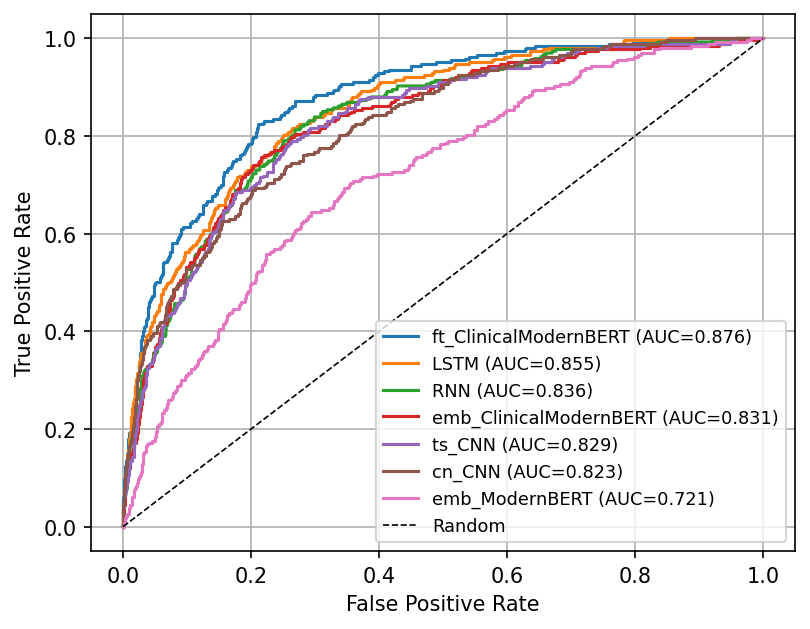}
    \caption{AUC Curve for all specialist classifiers.}
    \label{fig:auc_curves}
\end{figure}

\begin{figure}
    \vspace{1cm}
    \centering
    \includegraphics[width=0.5\linewidth]{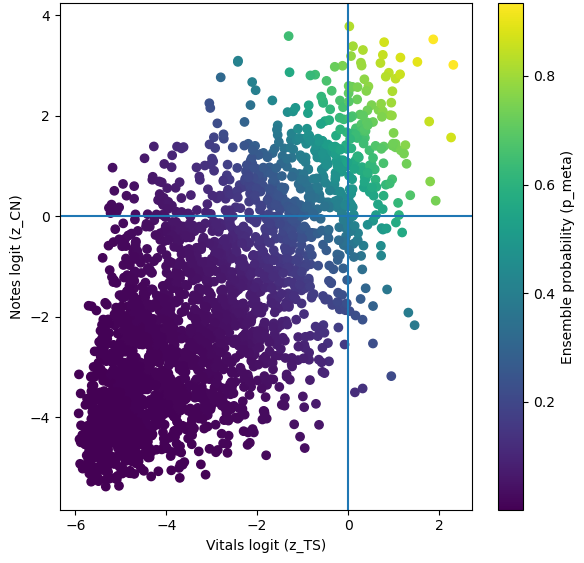}
    \caption{Complementarity scatter of base-model logits (log-odds), vitals $z_{\text{TS}}$ vs.\ notes $z_{\text{CN}}$.
    Vertical/horizontal lines at \(0\) correspond to each branch’s \(p=0.5\) threshold and partition agreement (↑↑/↓↓) and conflict (↑↓) quadrants.
    Point color encodes the ensemble probability \(p_{\text{meta}}\).}
    \label{fig:scatter}
\end{figure}

\begin{figure}[ht]
    \centering
    \includegraphics[width=0.5\linewidth]{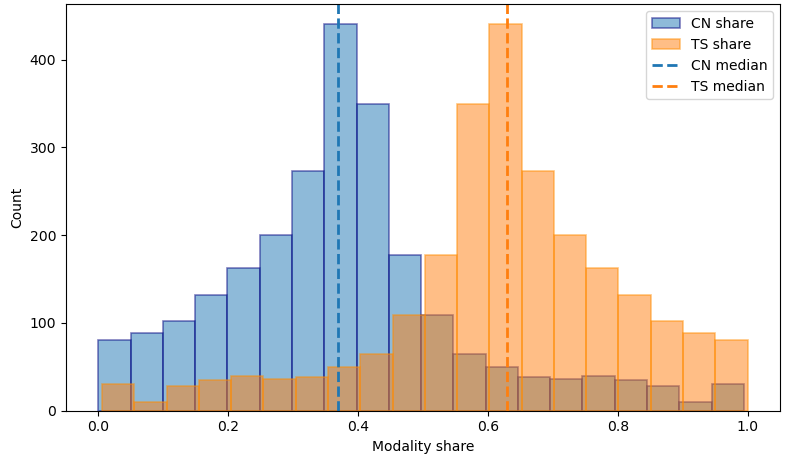}
    \caption{Distribution of modality shares on the test set. For each episode, the notes (CN) share is
    $|c_{\text{CN}}|/(|c_{\text{TS}}|+|c_{\text{CN}}|)$ and the vitals (TS) share is $1-\text{CN share}$.
    Dashed lines mark medians.}
    \label{fig:hist_of_share}
\end{figure}

\twocolumn

\end{document}